# Mathematical Formalized Problem Solving and Theorem Proving in Different Fields in Lean 4

Xichen Tang

## Abstract

Formalizing mathematical proofs using computerized verification languages like Lean 4 has the potential to significantly impact the field of mathematics, it offers prominent capabilities for advancing mathematical reasoning. However, existing efforts are largely limited to creating formalized versions of proofs from extensive online mathematical corpora, struggling to keep pace with the rapidly evolving nature of mathematics. To bridge the gap between traditional and computerized proof techniques, this paper explores the use of Large Language Models (LLMs) to generate formal proof steps and complete formalized proofs. By converting natural language (NL) mathematical proofs into formalized versions, this work introduces the basic structure and tactics of the Lean 4 language. The goal is to determine how AI can be leveraged to assist the mathematical formalization process and improve its performance. Several examples are provided that demonstrate solving problems using both traditional and Lean 4-based approaches. Ultimately, this paper presents an explanation of the foundations of Lean 4 and comparative analyses of the mathematical formalization process using traditional and AI-augmented techniques. The findings indicate that AI-powered tools have significant potential to accelerate and enhance the formalization of mathematical proofs, paving the way for more efficient and reliable theorem-proving for AI for Math in the future.

## 1   Introduction

Proving mathematical theorems is a core objective in the field, highlighting one's proficiency in logical reasoning and the capacity to navigate an infinite range of possibilities toward a solution. This endeavor demands intricate mathematical reasoning and exceptional problem-solving abilities. Consequently, enhancing theorem-proving capabilities has consistently been a central theme in artificial intelligence (AI) research. In recent years, the new paradigm of "AI for Science," which facilitates scientific research through artificial intelligence, has flourished. However, artificial intelligence still faces significant challenges in mathematics, particularly in the domain of theorem proving. Mathematical formalization and theorem proving have traditionally been the domain of human mathematicians, requiring substantial time and effort and carrying risks of human error and omission. The advancement of artificial intelligence presents new opportunities in this realm. Compared to language models like GPT-4, whose logical ability is inadequate for solving complex math problems, and symbolic AI systems like Sora, which may violate the laws of nature in mathematical modeling, the interactive theorem prover Lean 4 has emerged as an advanced tool for mathematical formalization[1]. Mathematicians have also launched their projects while for each there will be developments of many of the foundational theories of mathematical theorems and also provide lectures about how to organize a genuine large-scale formalization project. Moreover, these projects are sufficiently modular that all the participants can make meaningful contributions to the project without having to master all the mathematical prerequisites needed to understand the whole proof. The formalization projects are likely to be associated with other concurrent projects. For instance, one of the aims of the prime number theorem formalization project is to establish the Chebotarev density theorem, a version of which is needed at a certain step of the proof of FLT. And the proof of Katalin Marton's polynomial Freiman-Ruzsa conjecture [2], which was formalized in just three weeks, in which Tao used the Lean 4 blueprint software developed by Patrick Massot[3] to present the web page with a LaTeX/Lean 4 hybrid document, which contains an explanation of the proof that is comprehensible to both human and computer. In this article, I will introduce how Lean 4 works, the verification of its compilation process, and how AI can assist with mathematical formalization. I will also compare the use of Lean 4 to formalize proofs in various fields of mathematics, including Mathematical Olympiad problems in algebra, number theory, and geometry, as well as the Sylow theorems. This will shed light on the differences and advantages between the traditional mathematical proof approach and Lean 4 ones.

## 2 Basic structure and principles of the language of Lean 4

Before one starts from the mathematical formalization, one needs to be familiar with the basic structure and principles of the language of Lean 4. The computerized simulations have been enhanced by equipping artificial intelligence to solve mathematical problems, in which the AI part acts as a strong algorithm since it turns out to be quite promising in the Auto-Regressive models (ARMs) and the LLMs. Specifically, Large Language Models have shown remarkable abilities in a wide range of natural language processing tasks, particularly in addressing mathematical problems[5]. This makes the LLMs the fundamental logic structure model for AI in Lean 4. Specifically, in Lean 4, LLMs can substantially advance formal verification and theorem proving through several key mechanisms:

- Automated Proof Construction: LLMs can autonomously generate potential proof strategies or steps for given mathematical theorems or problems. Utilizing advanced natural language processing and mathematical reasoning capabilities, these models offer valuable insights and sequences of steps, thereby alleviating the manual effort traditionally involved in proof development.
- Code Synthesis and Optimization: These models can translate mathematical expressions and theorem requirements into Lean 4 code as they facilitate the creation of new definitions, lemmas, or theorem implementations and enhance the efficiency and structure of existing code through optimization techniques.
- Error Detection and Correction: LLMs can scrutinize Lean 4 code to identify logical errors, inconsistencies, or gaps in proofs. By leveraging their pattern recognition and inference capabilities, they provide recommendations for correcting and refining the code, thereby improving overall accuracy and reliability.
- Documentation and Explanation: LLMs are capable of generating comprehensive explanations and documentation for Lean 4 code and proofs. This includes elucidating proof strategies, providing context for mathematical concepts, and explaining code functionality, thus supporting users in understanding and applying formal verification tools more effectively.

- Interactive Guidance and Support: In the process of formal proof development, LLMs can act as interactive assistants, addressing user queries, offering additional explanations, and suggesting solutions to encountered challenges.

Mini-F2F (Minimal Functional First-Order Form) is a significant development in the context of Lean 4 and formal verification. It is a format designed to streamline and optimize logical expressions and proofs. In the domain of formal verification and automated theorem proving, the underlying principles of Mini-F2F involve several core concepts:

1. Minimal Functional Form

Minimal Functional Form represents a simplified version of logical expression that focuses on the most basic components. The aim is to reduce redundancy and complexity, making logical formulae more efficient to process. This emphasizes expressing logical statements with minimal syntax and structure, thereby reducing system resource requirements and enhancing speed.

2. First-Order Logic

First-Order Logic is a widely-used logical system that allows statements about individual objects and their relationships. Formulae in first-order logic include quantifiers (such as ∀ (universal quantifier) and ∃ (existential quantifier)), predicates, and functions.

In this context, Mini-F2F translates logical expressions into a standardized first-order logical form, facilitating effective handling by formal verification tools.

3. Functional Nature

Functional Nature underscores the computational and inferential capability of logical expressions, where each component of the formula has a clear purpose. In Mini-F2F, this functionality is reflected in simplifying logical expressions to their fundamental units, making reasoning and manipulation more straightforward and efficient.

4. Simplification and Optimization

Mini-F2F aims to simplify logical expressions, thereby reducing complexity and optimizing the proof process. By eliminating unnecessary structural elements and redundant information, it focuses on core logical relationships. Such simplification enhances the efficiency of processing logical expressions and verifying proofs, lowering computational resource demands.

5. Standardized Representation

Mini-F2F provides a standardized method for representing logical problems and proofs. This standardization improves interoperability and sharing of logical formulae across different formal verification systems. By adopting a unified representation, Mini-F2F reduces conversion complexity between systems, promoting compatibility of tools and methods.

6. Application in Formal Verification

In formal verification, the simplified form of Mini-F2F aids in the precise definition and verification of mathematical theorems and logical proofs. It offers a clear and concise way to represent logical expressions, supporting a more efficient verification and proving process.

**2.1 Lean 4 proving simple problem**

```
import Mathlib.Data.Real.Basic
example (a b c : R) : c * b * a = b * (a * c) := by
  rw [mul_comm c b]
  rw [mul_assoc b c a]
  rw [mul_comm c a]
```

▼ Tactic state
1 goal
a b c : R
⊢ b * (c * a) = b * (a * c)

*Figure 2.1.1 a simple proof of the commutative law and its associativity of multiplication by Lean 4, which can be simplified to basic steps that include the import of the package of the specific field of math in Lean 4 and give out the example(theorem) to prove and the just use the basic tactic and lemmas to rewrite.*

The Lean 4 theorem prover seeks to integrate interactive and automated theorem proving by combining automated tools and techniques with a framework that facilitates user interaction and the development of comprehensive axiomatic proofs. Its objective is to advance mathematical reasoning and the analysis of complex systems., ensuring the validation of statements across both fields. In Lean 4, users can directly utilize all internal data structures used in Lean 4 by importing the package/database, making it also a platform for developing effective domain-specific automation.

Starting from the comparison of Lean 4's proof and the natural language, we can start from a very basic example of a simple function that could be solved with only twice calculation.

```
6   --generated proof
7   theorem mathd_algebra_270
8   (f : R → R)
9   (h0 :∀x,x=-5→fx=1/(x+5)):
10  f (f 1) = 6/31 := by
11  -- We see that f 1 = 1 / (1 + 5) = 1 / 6
12  haveh1 :f1=1/6:=bynorm_num[h0]
13  -- Thus f (f 1) = f (1 / 6) = 1 / (1 / 6 + 5) = 1 / (31 / 6) = 6 / 31 calc
14  f (f 1) = f (1 / 6) := by rw [h1]
15  _ = 1 / (1 / 6 + 5) := by norm_num [h0]  _ = 1 / (31 / 6) := by norm_num
16  _ = 6/31 := by norm_num
```

If $f(x) = \frac{1}{x+5}$, what is $f(f(1))$? Show that it is $\frac{6}{31}$.
**Proof:**
First, calculate $f(1)$:
$$f(1) = \frac{1}{1+5} = \frac{1}{6}$$
Now, calculate $f(f(1)) = f\left(\frac{1}{6}\right)$:
$$f\left(\frac{1}{6}\right) = \frac{1}{\frac{1}{6}+5} = \frac{1}{\frac{1}{6}+\frac{30}{6}} = \frac{1}{\frac{31}{6}} = \frac{6}{31}$$
Thus, $f(f(1)) = \frac{6}{31}$.

*Figure 2.1.2: a simple problem as an easy attempt in both natural language and Lean 4*

The basic method of proving mathematics in Lean 4 is to do it step by step, if the proof is too complicated, we can separate them into parts individually and combine them at the end of the proof later, specifically in complicated proofs of theorems, etc.

**2.2.1 The Operational Logic and Underlying Principles with Mathematical Basis of Lean 4**

To produce fully formalized and rigorous proofs, we need methods to represent mathematical objects. This necessitates a discussion of the foundations of mathematics, which allows us to construct and discuss the myriad mathematical objects[4]. A logical framework is required in which we can discuss formal mathematical theories. The system itself specifies certain deductive methods known as inference rules. Often, we also need to introduce additional statements within the system, called axioms. When discussing the foundations of mathematics, set theory is used to characterize sets and their membership relations through axioms. By encoding various mathematical objects as sets and performing operations on sets to manipulate mathematical objects, mathematical theories are developed.

For instance, *ZFC set theory*, which is based on first-order logic, uses several axioms to characterize sets within first-order logic. Type theory allows for the definition of types, which are used to obtain and work with values within those types, and often supports some simple automated computations. Type theory itself is a reasoning system with inference rules for type operations, and it usually requires fewer additional axioms. There are extra rules to assist with automatic simplifications, type judgments, and equality checks. So why choose type theory over set theory, set theory lacks type checking; all objects are sets, which allows for unexpected statements like discussing what $\pi \cap sin$ is, or whether $R \in 0$ is correct. Under some common encoding methods, it can be proven that $2 = (0,0)$ and that this is a topology on 1. Most mathematicians do not view mathematical objects as encoded sets but think more in terms of type theory, defining objects by their inherent properties. Type theory is naturally suited to describe computational processes and can automatically simplify results, whereas functions in set theory are special ordered pairs of sets. Just as there are many different axiom systems in set theory, there are also numerous variants of type theory. Many of these are sufficiently strong to serve as a foundation for mathematics. For example, one can construct a model of ZFC in Lean 4. Type theory provides a powerful logical system and also serves as a mathematical model for the type systems of programming languages, guiding language design. The following introduces common concepts in type theory and how they are used as logical systems:

Each type has certain values, and the type of an expression can also be determined. When you want to explicitly mark the type of an element $a$ as $A$, you can write $a:A$. This is analogous to set theory, where we write $0 \in N$ or $\pi \in R$. However, in set theory, $\in$ denotes a relation that discusses whether an object belongs to a set, whereas in type theory, : is a judgment at the meta-language level. This is similar to general programming languages; for example, in C++ there are types such as $int, double, and\ std::cin : std::istream$.

In type theory, the focus is on two main questions: how to obtain a value of a type and how to use a value of a type. The method for obtaining a value of a type is known as construction rules, while the method for using a value of a type is known as elimination rules. The following sections will introduce some types in type theory in Lean 4.

For types A and B, we have the function type $A \to B$. If for an element $a:A$, there exists an expression $\Phi:B$ containing $a$, we can define a function $f(a) := \Phi$, with $f:A \to B$. For example, for $n:N$, there is a function $f(n) := n + n$, hence $f:N \to N$. Sometimes, this function is written as $\lambda n. n + n$, analogous to the notation $n \mapsto n + n$ used by mathematicians. The elimination rule for functions is function application: for $f:A \to B$ and $a:A$, we write the application as $fa:B$. The function type is fundamentally important; modern type theory began with Church's introduction of the simply typed lambda calculus in 1940, which is based on the concept of function abstraction.

For types A and B, we have the product type $A \times B$. In set theory, this is analogous to the set of all pairs $(a, b)$ where $a \in A\ and\ b \in B$. The construction rule is $A \to (B \to A \times B)$. The elimination rules are $fst: A \times B \to A$ and $snd: A \times B \to B$. If types are viewed as objects in a category and functions as morphisms, then $A \times B$ is the product object of $A$ and $B$.

For types A and B, we have the sum type $A + B$ (which might be denoted differently in some sources). Analogous to set theory, this represents the disjoint union of $A\ and\ B$, where all pairs $(0, a)\ for\ a \in A\ and\ (1, b)\ for\ b \in B\ form\ A \sqcup B$. The construction rules are $inl: A \to A + B$ and $inr: B \to A + B$. The elimination rule is $(A \to C) \to (B \to C) \to (A + B \to C)$. We immediately see that $A + B$ is the coproduct object of $A\ and\ B$, and it is also called the *coproduct type*.

For the empty type, which has no values, the notation 0 is used. There are no construction rules for this type. The elimination rule is to show that any type can be derived from the empty type. Possessing a value of the empty type is absurd. In category theory, the empty type corresponds to the initial object in a category. The terminal object in a category is a single-element type, denoted 1 here.

In Lean 4, the $\Pi$ type is a fundamental type, with function types being a special case of it. The other types introduced above are special cases of inductive types. The construction rules involve specified functions and can include self-referential inputs. The elimination rules involve "induction" or "recursion," ensuring that each value of an inductive type in an empty context is a tree of expressions constructed from the rules. Examples include natural numbers, lists, and binary trees.

The category formed by the types together with the specified construction functions forms a category where inductive types are the initial objects. There are many other examples, and all these are the basic and type of math for basis of Lean 4. Taking implication as an example, if we have a proof of $p \to q$ and a proof of $p$, then we can obtain a proof of $q$. For conjunction, if we have proofs of p and q, we can derive a proof of $p \wedge q$. For disjunction, if we have a proof of $p$, we can derive a proof of $p \vee q$, and similarly for $q$. Given a proof of $\bot$, by the principle of explosion, we can derive a proof of any proposition. Analogous to the construction rules for function types, product types, and sum types, as well as the elimination rule for empty types, propositions can be viewed as types and proofs as programs in type theory. Verifying the type of a program is equivalent to verifying the correctness of the proof. This correspondence between type systems and natural deduction in logic is known as the Curry-Howard correspondence. Regarding negation, it can be defined as implication to falsehood, which in type theory corresponds to $P \to 0$. The rationality of this can be explored from the perspective of force or equivalence to falsehood. The focus is not on how a proposition is proven. Any proof (i.e., value) of a proposition should be considered equal. In Lean 4, types in *Prop* are viewed as propositions. Unlike other types, propositions are proof-agnostic. Different proofs of a proposition are definitionally equal and indistinguishable, so users need not concern themselves with how a proposition is proven. Hence, unlike other types, a proposition is either empty or has a single value. Further details of type theory, such as universe levels and axioms, can be found in *The Type Theory of Lean 4*. Our understanding of mathematical objects evolves at different stages, encompassing structures such as order, algebra, topology, and measure on integers and real numbers. These structures might also satisfy additional properties. When discussing a specific mathematical object, we often assume it has some additional structure without always specifying what. We need to support adding default structures to mathematical objects and discussing them in terms of these structures. In proof assistants like Lean 4, this is often represented by type classes. Type classes provide additional information related to certain types or values and can be globally defined or scoped locally, with Lean 4 automatically resolving instances as needed. Type class systems involve more than simple lookups; they require search and synthesis capabilities, such as ordering, algebraic, and topological structures on product types. Type classes also facilitate other tasks, such as type conversion, with potential for further simplification and algorithmic development. For more on this, see *Theorem Proving in Lean 4*[13]. Inductive relations, such as those from metric spaces to topological spaces, are addressed in mathlib with default definitions for metric spaces, though manual supplementation is possible. Additionally, Lean 4's cloud programming (syntax and expressions) greatly contributes to its clarity and precision, allowing it to function as a general functional programming language. Lean 4 supports remote execution of code, enabling collaborative development of proofs and programs. This means that users can work on the same Lean 4 project from different locations, with changes being synchronized in real-time[7]. Tools like Lean 4's web-based editor and proof assistants allow users to write and test Lean 4 code directly in a browser. These environments often provide a shared platform where users can interact with Lean 4's features without needing a local installation. Tactics and Meta-programming: Lean 4 includes a robust metaprogramming framework, which allows users to write custom tactics and automation scripts that interact with the Lean 4 environment. This capability is particularly useful for creating advanced proof strategies and automating repetitive tasks. Lean 4's type class system, combined with metaprogramming, enables dynamic resolution of type class instances. This means that Lean 4 can automatically infer and apply appropriate instances based on the context of the proofs and programs being developed. Lean 4's ecosystem includes extensive libraries like mathlib, which are accessible both locally and through cloud-based environments. These libraries provide a wealth of mathematical knowledge and tools that can be used and extended in Lean 4 projects. Lean 4's core strength lies in its ability to verify proofs and ensure correctness. Its cloud programming capabilities facilitate the efficient execution and verification of complex proofs by harnessing remote computational resources. Lean 4's support for functional programming paradigms and advanced type systems aids in the development of efficient algorithms and data structures. Metaprogramming allows for the creation of optimized solutions while maintaining rigorous correctness guarantees[6]. The Lean 4 compiler is primarily implemented in Lean 4 itself, supporting metaprogramming and tactic writing. Lean 4 verifies the expressions or "certificates" computed by programs, and metaprograms themselves are not required to have termination or correctness[12].

**2.2.2 The Compiling process of Lean 4**

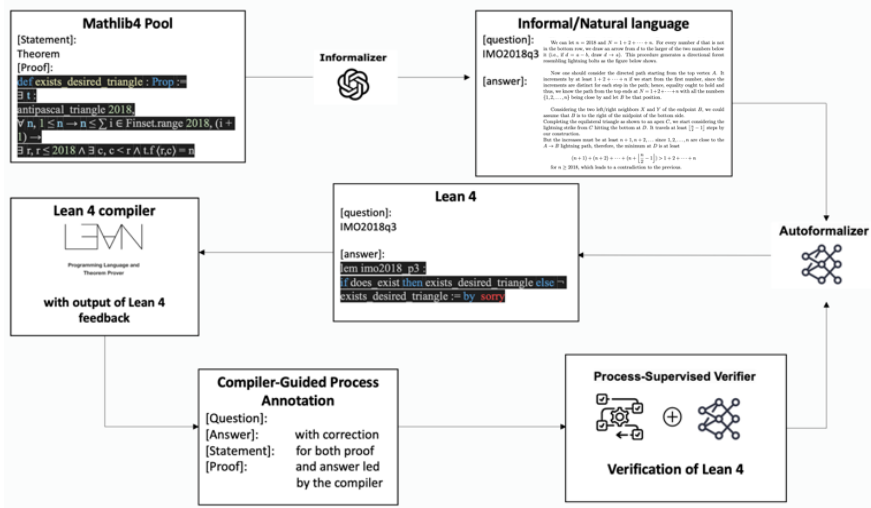

*Figure.2.2.2.1: the basic steps of Lean 4's proof*

As the graph above shows the computerized and compiling logic, computers primarily assist in establishing a claim through two main methods: they can aid in initially discovering a proof, and they can also help verify the accuracy of a claimed proof. Automated theorem proving focuses primarily on discovering proofs and offers methods for validating the correctness of statements in propositional and first-order logic. The increase in complexity of computation has also led to a huge shift in the choice of mathematical formal theorem-proving methods, which is extremely complex if AI is to be able to automatically prove mathematical theorems, such as probabilistic reasoning patterns. Unlike other methods, interactive theorem proving focuses more on 'verification,' demanding that each assertion is backed by a proof within a suitable axiomatic framework. This imposes a stricter criterion: each rule of reasoning and each calculation step must be validated by referring back to earlier definitions and theorems, ultimately tracing back to fundamental axioms and rules. Many of these systems offer detailed "proof objects" that can be exported and independently verified by other systems. While creating these proofs often demands more user input and interaction, it allows for the formalization of more complex and significant proofs.

## 3 The assistance of AI for mathematical formalization

### 3.1 The development and application of synthetic training data in Lean 4

To address the situation of lacking the scarce training data available to train the AI in the formal language of Lean 4, mathematicians have suggested a new process that sequentially creates and refines artificial data to translate mathematical problems in natural language into Lean 4 statements, and vice versa. This shows the fact that the synthetic training data can offer useful training examples and improve the performance of LLMs[10] in translating and understanding complex mathematical problems and proofs to do formalization, which means AI can assist mathematical formalization. If one feels mediocre about one's computerized formalization, one should rather do mathematical formalization manually. To avoid lengthy proofs, massive training of the AI is valuable since it does not know whether the proof it is correct or not, and consequently people need to input more details. With these data, I assume, can effectively improve the reasoning ability of this AI[11]. One can use machine learning to fit a function and construct and solve the hypothesis. An obvious question when formalizing a math paper is the degree of change that the proof needs to be changed to be formalized: in other words, whether there were any flaws in the paper. The majority of the errors I identified were easily rectifiable without compromising the essence of the argument, so it can be reasonably argued that the Lean 4 version of the proof is essentially a translation of the informal proof. Lean 4's underlying logic can be interpreted computationally, much like a programming code. Moreover, it could be seen as a system for writing programs with accurate semantics, as well as formalizing and reasoning about the mapping process that the programs compute. It can also be viewed as an extensible theorem prover and an efficient programming language which needs natural language annotation and a value judgment for the computer to understand the idea to give good tactics and lemmas. One advantage that one can probably draw from applying Lean 4 to verify one's proof is correct or not is that Lean 4 has a basic model of inference with logic involved. The other is a basic model for automatic reasoning with another one for computation. Formal verification utilizes logical and computational approaches to substantiate assertions articulated in precise mathematical language. In practical terms, the distinction between verifying a mathematical proposition and ensuring the accuracy of a system becomes unclear: formal verification requires representing hardware and software systems mathematically, thereby making the process of confirming their correctness similar to proving theorems. On the contrary, demonstrating a mathematical theorem could involve extensive computation.

### 3.2 The verification of compiling state in Lean 4

In Lean 4, ensuring that your formalized theorems are correct relies on Lean 4's built-in proof assistant and formal verification mechanisms. In the Lean 4 environment, we write and run a code. Lean 4 will attempt to automatically deduce and verify the proof. If one's code is correct, Lean 4 will confirm the proof is complete without errors or warnings. It provides powerful tools and tactics to help you build and verify proofs such as tactics:

use Lean 4 tactics (e.g. *rw*, *apply*, *exact*) to automate the proof process which can help you decompose problems and build proofs step by step, interactive proofs since one can interactively step through the proof process, with Lean 4 providing feedback on each step to help you correct mistakes and adjust our proof strategy. Besides, in Lean 4, one can use the *check* or *#check* command to verify the type and construction of the theorem. This command will display the type of *my_theorem*, helping you ensure it meets expectations. As Lean 4's built-in verification mechanism ensures that one's provided proof is logically sound and conforms to the definitions, compiling and verifying the Lean 4 files will catch any formal errors.

There is also type system verifies that your theorem and proof adhere to the defined types. Any type mismatches or proof errors are detected during compilation. Ensuring the correctness of a theorem involves not just syntactic and structural correctness but also the logical validity of formalized definitions and proofs, Lean 4's formal verification system ensures that the theorem one's written is mathematically correct.

## 4 Lean 4 proving in Mathematical Olympiad (MO)

To demonstrate whether using Lean 4 is better than the traditional written proof in natural language, I will compare the solutions for questions in the mathematical Olympiad in different the four topics of MO and the computerized proof by Lean 4[9].

**4.1 Using NL (natural language) and**

**Lean 4 to prove algebraic questions in MO**

The first sample topic I chose is the $6^{th}$ problem of IMO 2024, which is an algebra question in MO. Here is the proof written in natural language:

**The IMO 2024 Question 6:**
Let $Q$ be the set of rational numbers. A function $f : Q \to Q$ is called aquaesulian if the following property holds: for every $x, y \in Q$,

$$f(x + f(y)) = f(x) + y \quad \text{or} \quad f(f(x) + y) = x + f(y).$$

Show that there exists an integer $c$ such that for any aquaesulian function $f$, there are at most $c$ different rational numbers of the form $f(r) + f(-r)$ for some rational number $r$, and find the smallest possible value of $c$.

**Solution:**
The smallest value is $c = 2$. Supposing that $f$ is a function satisfying the problem cases, we know:

- $a \approx b$ if either $f(a) = b$ or $f(b) = a$,
- $a \to b$ if $f(a) = b$,
- $P(x, y)$ to denote the proposition that either $f(x+f(y)) = f(x)+y$ or $f(f(x)+y) = x + f(y)$,
- $g(x) = f(x) + f(-x)$.

Hence, the condition $P(x, y)$ could be rephrased as saying that $x + f(y) \approx f(x) + y$, and we ought to determine the maximum possible number of elements of $\{g(x) \mid x \in \mathbb{Q}\}$.

Providing an example of a function $f$ for which there are two values of $g(x)$, we take the function $f(x) = \lfloor x \rfloor + \{x\}$, where $\lfloor x \rfloor$ denotes the floor of $x$ (that is, the largest integer less than or equal to $x$) and $\{x\} = x - \lfloor x \rfloor$ denotes the fractional part of $x$.

Firstly, one should show that $f$ satisfies $P(x, y)$. Given $x, y \in \mathbb{Q}$, one have

$$f(x) + y = \lfloor x \rfloor + \{x\} + \lfloor y \rfloor + \{y\} = (\lfloor x \rfloor + \lfloor y \rfloor) + (\{x\} + \{y\}),$$

$$x + f(y) = \lfloor x \rfloor + \{x\} + \lfloor y \rfloor + \{y\} = (\lfloor x \rfloor + \lfloor y \rfloor) + (\{x\} + \{y\}).$$

If $\lfloor x \rfloor < \lfloor y \rfloor$, one we have that the fractional part of $f(x)+y$ is $\lfloor y \rfloor - \lfloor x \rfloor$ and the floor is $\lfloor x \rfloor + \lfloor y \rfloor$, so $f(x)+y \approx x+f(y)$. Likewise, if $\lfloor x \rfloor > \lfloor y \rfloor$, then $x+f(y) \approx f(x)+y$. Finally, if $\lfloor x \rfloor = \lfloor y \rfloor$, then $f(x)+y = x+f(y) = \lfloor x \rfloor + \lfloor y \rfloor$ is an integer. In all cases, the relation $P$ is satisfied.

Lastly, one can notice that if $x$ is an integer then $g(x) = 0$, otherwise $g(x) = -2$, so there are two values for $g(x)$ as required.

Now, we prove that there cannot be more than two values of $g(x)$. $P(x, x)$ tells us that $x + f(x) \approx x + f(x)$. So for all $x$,

$$f(x + f(x)) = x + f(x). \tag{1}$$

we should begin with the following lemma.

**Lemma** $f$ is a bijection, and satisfies

$$f(f(f(x))) = x. \tag{2}$$

**Proof.** I can first prove that $f$ is injective by supposing that $f(x_1) = f(x_2)$; then $P(x_1, x_2)$ tells us that $f(x_1) + x_2 \approx f(x_2) + x_1$. Without loss of generality, I suppose that $f(x_1) + x_2 \approx f(x_2) + x_1$. But $f(x_1) = f(x_2)$, so $f(f(x_1) + x_2) = f(f(x_2) + x_2) = f(x_2) + x_2$ by (1). Therefore,

$$f(x_2) + x_1 = f(x_2) + x_2,$$

as required.

Now, (1) with $x = 0$ tells us that $f(f(0)) = f(0)$ and so by injectivity $f(0) = 0$. By applying $P(x, -f(x))$, we know that $0 \approx x + f(-f(x))$, so either $0 = f(0) = x + f(-f(x))$ or $f(x + f(-f(x))) = 0$ which implies that $x + f(-f(x)) = 0$ by injectivity. Either way, one can deduce that $x = -f(-f(x))$, or $x = f(-f(x))$ by replacing $x$ with $-x$.

Finally, note that bijectivity follows immediately from (2).

Since $f$ is bijective, it has an inverse, which I denote $f^{-1}$. Rearranging (2) (after replacing $x$ with $-x$) gives that $f(-x) = -f^{-1}(x)$. One know has $g(x) = f(x) + f(-x) = f(x) - f^{-1}(x)$.

Suppose $g(x) = u$ and $g(y) = v$, where $u \neq v$ are both nonzero. Define $x_1 = f^{-1}(x)$ and $y_1 = f^{-1}(y)$; by definition, we have

$$x_1 \approx x \approx x_1 + u,$$

$$y_1 \approx y \approx y_1 + v.$$

Putting in $P(x_1, y)$ gives $x + y \approx x_1 + y_1 + v$, and putting in $P(x, y_1)$ gives $x + y \approx x_1 + y_1 + u$. These are not equal since $u \neq v$, and $x + y$ may have only one incoming and outgoing arrow because $f$ is a bijection, so we must have either $x_1 + y_1 + u \approx x + y \approx x_1 + y_1 + v$ or the same with the arrows reversed. Swapping $(x, u)$ and $(y, v)$ If needed, we can assume, without loss of generality, that the arrows are directed correctly.

Also, one has $-x_1 - u \approx -x \approx -x_1$ by the Lemma. Putting in $P(x+y, -x_1-u)$ gives $y \approx y_1 + v - u$, and so $y_1 + v - u$ must be either $y_1 + v$ or $y_1$. This means $u$ must be either 0 or $v$, and this contradicts our assumption about $u$ and $v$.

We can see that this problem uses basic abstract algebra knowledge, this solution is inspired by the IMO-official solution combining my own one. Starting by proving that $f$ must be surjective. Then we can suppose it is not, there should be some $t$ that does not appear in the output of $f$. After which $P(x, t - f(x))$ tells us about the relationship between $t$ and $f$, and so by assuming $f(t) = x + f(t - f(x))$ for all $x$, which leads to a contradiction to our assumption before. Hence, one finds the value of $t$ when $f(t) = 0$; such $t$ must exist by surjectivity. Therefore, we can solve the problem. Utilizing the same idea, now we can start the formalization of the Lean 4 version of this problem:

*Figure.4.1.1: After importing the database, define the variables and the properties, use different lemmas and tactics to formalize.*

Then we can continue the general proof keep using the lemmas and tactics we need for each individual part of the formalization.

*Figure.4.1.2: continued formalization*

Continue using the same idea, we find the crucial insight is that $f(-f(-x)) = x$. By examining a pair of distinct nonzero values of $f(x) + f(-x)$ and performing a series of manipulations in conjunction with this insight, we arrive at a contradiction. Consequently, there are at most two distinct values of $f(x) + f(-x)$. This analysis is valid in any *AddCommGroup*; specifically, over $\mathbb{Q}$, we demonstrate that $\lfloor x \rfloor - Int.fract\ x$ can take on exactly two different values of $f(x) + f(-x)$. And we could finish the proof. We consider the individual property of those as we use the *abel* processing tactic and finish the proof. In this sample using Lean 4 to prove this algebra question turns out to be barely satisfying since it some process still needs to be done by me to lead it to a right direction.

**4.2 The use of NL and Lean 4 to prove the geometry problems in mathematical Olympiad**

Another sample I chose is to prove the geometric questions in MO which is the 2$^{nd}$ IMO question in 2019. Here is the question and the natural language solution written by LaTeX:

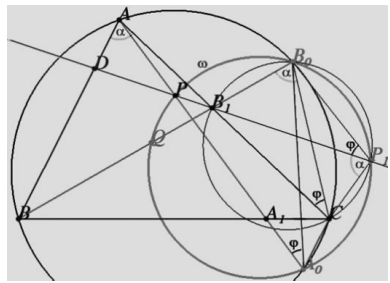 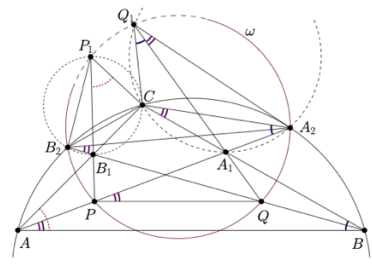

*Figure.4.2.1&4.2.2: the 1$^{st}$ graphs drawn(idea of the graph is inspired by the AOPS [8])     2$^{nd}$ graph given by the IMO official solution is given on official website ([www.imo2019.uk](www.imo2019.uk))*

**The IMO 2019 Question 2:**
In triangle $ABC$, point $A_1$ lies on side $BC$ and point $B_1$ lies on side $AC$. Let $P$ and $Q$ be points on segments $AA_1$ and $BB_1$, respectively, such that $PQ$ is parallel to $AB$. Let $P_1$ be a point on line $PB_1$, such that $B_1$ lies strictly between $P$ and $P_1$, and $\angle PP_1C = \angle BAC$. Similarly, let $Q_1$ be a point on line $QA_1$, such that $A_1$ lies strictly between $Q$ and $Q_1$, and $\angle CQ_1Q = \angle CBA$.

Prove that points $P, Q, P_1$, and $Q_1$ are concyclic.

**Proof:**
Given triangle $ABC$, points $A_1$ on $BC$ and $B_1$ on $AC$. Points $P$ and $Q$ on segments $AA_1$ and $BB_1$ such that $PQ \parallel AB$. Points $P_1$ on line $PB_1$ with $B_1$ between $P$ and $P_1$, and $\angle PP_1C = \angle BAC$. Similarly, points $Q_1$ on line $QA_1$ with $A_1$ between $Q$ and $Q_1$, and $\angle CQ_1Q = \angle CBA$. We need to prove that $P, Q, P_1, Q_1$ are concyclic. The first and the most important part of the proof is to construct a circle passing through points $P, Q, A_0$, and $B_0$, and demonstrate that points $P_1$ and $Q_1$ lie on this circle.

Assuming that the intersection point of lines $AP$ and $BQ$ lies on segment $PA_1$, if it lies on segment $AP$, the proof remains the same, with adjustments to angles up to $180°$.

Let $\Omega$ denote the circumcircle of $\triangle ABC$. Points $A_0$ and $B_0$ are intersections of lines $AP$ and $BQ$ with $\Omega$. Suppose $\angle BAP = \delta$. Since $PQ \parallel AB$, it follows that $\angle QPA_0 = \delta$.

$\angle BAP = \angle BB_0A_0 = \delta$, as they intercept arc $BA_0$ of circle $\Omega$.
$\angle QPA_0 = \angle QB_0A_0 \Longrightarrow QPB_0A_0$ is cyclic (in circle $\Omega$).
Let $\angle BAC = \alpha$ and $\angle AA_0B_0 = \varphi$.
$\angle PP_1C = \alpha$, $\angle BB_0C = \alpha$ since they intercept arc $BC$ of circle $\Omega$. Thus, $B_0P_1CB_1$ is cyclic.
$\angle ACB_0 = \angle AA_0B_0 = \varphi$ (intercepting arc $A_0B_0$ of circle $\Omega$).
$\angle B_1CB_0 = \varphi$. $\angle B_1P_1B_0 = \angle B_1CB_0 = \varphi$ (intercepting arc $B_1B_0$ of circle $B_0P_1CB_1$).

Therefore, $\angle PA_0B_0 = \angle PP_1B_0 = \varphi$, indicating point $P_1$ lies on circle $\omega$. Similarly, point $Q_1$ also lies on circle $\omega$. Hence the points $P, Q, P_1$, and $Q_1$ are concyclic is proved.

In fact, one shall notice that there is also another approach given by the official. I would start with the Lean 4 version of the mathematical proof of this problem.

```
2019IMOq2.lean > ...
1  ∨ import MIL.Common
2    import Mathlib.Topology.Instances.Real
3    import Mathlib.Analysis.NormedSpace.BanachSteinhaus
4    import Mathlib.Geometry.Euclidean.Angle.Sphere
5    import Mathlib.Geometry.Euclidean.Sphere.SecondInter
6  ∨
7  ∨ open Affine Affine.Simplex EuclideanGeometry FiniteDimensional
8
9    open scoped Affine EuclideanGeometry Real
0
1    attribute [local instance] FiniteDimensional.of_fact_finrank_eq_two
2
3    variable (V : Type*) (Pt : Type*)
4
5    variable [NormedAddCommGroup V] [InnerProductSpace ℝ V] [MetricSpace Pt]
6
7    variable [NormedAddTorsor V Pt] [hd2 : Fact (finrank ℝ V = 2)]
8
9  ∨ namespace Imo2019P2
```

*Figure.4.2.3: A configuration that meets the criteria of the problem. One should start from the header files, it is worth noticing that I have changed some of the data base that I am using to prove this geometry problem after which one shall move to prove the small, individual part of the total proof.*

One establishes this framework to avoid the proliferation of hypotheses, which would otherwise hinder the required proofs. By importing the database and defining the variables related to the configuration, the eventual outcome regarding a problem statement that does not utilize this framework is then inferred from one formulated within this framework.

```
structure Imo2019q2Cfg where
  (A B C A₁ B₁ P Q P₁ Q₁ : Pt)
  affineIndependent_ABC : AffineIndependent ℝ ![A, B, C]
  wbtw_B_A₁_C : Wbtw ℝ B A₁ C
  wbtw_A_B₁_C : Wbtw ℝ A B₁ C
  wbtw_A_P_A₁ : Wbtw ℝ A P A₁
  wbtw_B_Q_B₁ : Wbtw ℝ B Q B₁
  PQ_parallel_AB : line[ℝ, P, Q] ∥ line[ℝ, A, B]
```

*Figure.4.2.4: a figure of the starting structure*

After assuming the implicit properties of the designated line and the initial designated angle, a default orientation must be chosen from the available lemmas. I define the variable and begin to observe that the configuration exhibits symmetry, enabling results proven for one point to be extended to apply to another.

```
 99    def symm : Imo2019q2Cfg V Pt where
100      A := cfg.B
101      B := cfg.A
102      C := cfg.C
103      A₁ := cfg.B₁
104      B₁ := cfg.A₁
105      P := cfg.Q
106      Q := cfg.P
107      P₁ := cfg.Q₁
108      Q₁ := cfg.P₁
109      affineIndependent_ABC := by
110        rw [← affineIndependent_equiv (Equiv.swap (0 : Fin 3) 1)]
111        convert cfg.affineIndependent_ABC using 1
112        ext x
113        fin_cases x <;> rfl
114      wbtw_B_A₁_C := cfg.wbtw_A_B₁_C
115      wbtw_A_B₁_C := cfg.wbtw_B_A₁_C
116      wbtw_A_P_A₁ := cfg.wbtw_B_Q_B₁
117      wbtw_B_Q_B₁ := cfg.wbtw_A_P_A₁
118      PQ_parallel_AB := Set.pair_comm cfg.P cfg.Q ▸ Set.pair_comm cfg.A cfg.B ▸ cfg.PQ_parallel_AB
119      P_ne_Q := cfg.P_ne_Q.symm
120      sbtw_P_B₁_P₁ := cfg.sbtw_Q_A₁_Q₁
121      angle_PP₁C_eq_angle_BAC :=
122        angle_comm cfg.C cfg.Q₁ cfg.Q ▸ angle_comm cfg.C cfg.B cfg.A ▸ cfg.angle_CQ₁Q_eq_angle_CBA
123      C_ne_P₁ := cfg.C_ne_Q₁
124      sbtw_Q_A₁_Q₁ := cfg.sbtw_P_B₁_P₁
125      angle_CQ₁Q_eq_angle_CBA :=
126        angle_comm cfg.P cfg.P₁ cfg.C ▸ angle_comm cfg.B cfg.A cfg.C ▸ cfg.angle_PP₁C_eq_angle_BAC
127      C_ne_Q₁ := cfg.C_ne_P₁
```

*Figure 4.2.5: a figure of the continuing proof using the symmetry of the configuration (with some tactics mentioned before)*

```
theorem A_ne_B : cfg.A ≠ cfg.B :=
  cfg.affineIndependent_ABC.injective.ne (by decide : (0 : Fin 3) ≠ 1)
```

*Figure.4.2.6: continued proof*

As the figure shown, properties of the configuration that are evident from the diagram depicting the circle, and construction of the points $A_2$ and $B_2$ should be done. After utilizing this approach, one shall keep constructing this structure similarly with more obvious configuration properties appear.

*Figure.4.2.7: continued proof relating the angles in the configurations and other properties of the patterns*

*Figure.4.2.8: continued proof using more properties of the geometric patterns & start angle chasing for more than three loops*

Allowing rays $AA_1$ and $BB_1$ to intersect the circumcircle of $ABC$ at points $A_2$ and $B_2$ respectively. After a series of angle-chasing steps, it can be demonstrated that $P, Q, A_2$, and $B_2$ lie on the same circle. Next, let ω denote the circle passing through these points to establish that $C, Q_1, A_2, and\ A_1$ are also concyclic, with $Q_1$ consequently lying on $\omega$. Similarly, $P_1$ lies on $\omega$, thus confirming that the four required points are concyclic.

*Figure.4.2.9: the continued proof and the ideal output shown on the Lean 4 info view, with the variables I defined at the beginning of the code*

One should observe that the majority of the formal proof concerns establishing non-degeneracy conditions essential for the angle-chasing and concyclicity arguments, whereas an informal solution omits discussion of these conditions altogether. Additionally, according to "*Geometry.Euclidean.Angle.Oriented.Basic*" tactics, oriented angles are considered $modulo\ 2\pi$, so aspects of the angle chase valid only for angles modulo $\pi$ (as utilized in the informal solution) are expressed as double-angle equalities. I applied the following conventions in formalizing IMO geometry problems: Unless explicitly stated otherwise, a problem is assumed to occur in a plane, obviating the need to prove coplanarity from other conditions. Angle references in problem statements are assumed to be unoriented. An angle $\angle XYZ$ implies that $X$ and $Z$ are distinct from $Y$, as irrelevant values do not affect informal mathematics, and these implications are treated as problem hypotheses if they derive from other hypotheses. Similarly, a line reference $XY$ implies $X \neq Y$ and is included as a hypothesis; if $XY$ is stated to be parallel to something, it is understood to be parallel as a line. Implicit assumptions regarding point distinctness are considered once per pair of points, even if inferred from multiple references to lines or angles. If $X \neq Y$ is given, it is not separately stated that $Y \neq X$. Such assumptions are omitted if the problem involves a triangle containing these points or specifies strict betweenness among three points including them, with betweenness understood to be strict. Segments and sides are presumed to include their endpoints unless this would invalidate the problem. Although degenerate cases may not have been explicitly considered when formulating the problem, contestants might not be expected to address them. A point's position on a side or segment is indicated directly using "*Wbtw*" rather than the more literal "*affineSegment*". About the tactics and lemmas that one is not familiar with, one can move the cursor upon the tactic/lemma and search for its definition and declaration in the compiler as the figure below shows (the figures are captured by the compiler *Visual Studio Code by Microsoft,* who also define the tactics and lemmas and give the idiographic definitions in Lean 4).

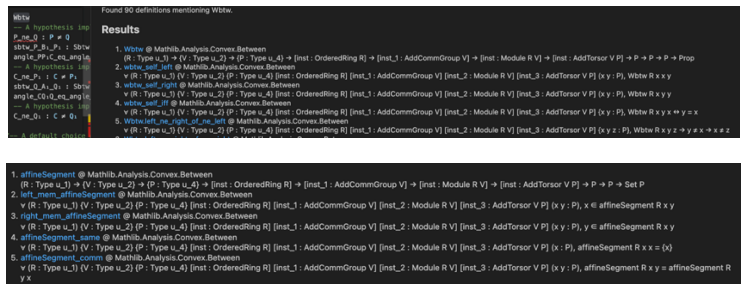

*Figure 4.2.10&11: The related definition given by the database mathlib for the two tactics mentioned*

Using Lean 4 for geometric problems in MO appears to be quite promising for the reason that the proposal of using *Euclidean plane geometry* could eliminate the need for demonstrations led by mathematicians by employing various tactics and lemmas while synthesizing and integrating millions of theorems and proofs of varying complexity levels, similar to the process of *the Google Deepmind,* but with the LLMs instead of *AlphaGeometry's* neuro-symbolic system which uses a neural language model .

**4.3 The use of NL and Lean 4 to prove the number theory topic in mathematical Olympiad**
The 1st sample for number theory I chose is the 2nd IMO question in 2024, which could be found on the IMO official website. Here is the written solution of the problem:

**The IMO 2024 Question 2:**
Determine all pairs $(a, b)$ of positive integers for which there exist positive integers $g$ and $N$ such that
$$\gcd(a^n + b, b^n + a) = g$$
holds for all integers $n \geq N$. (Note that $\gcd(x, y)$ denotes the greatest common divisor of integers $x$ and $y$.)

**Solution:**
**Lemma:** We have that $g = \gcd(a, b)$ or $g = 2\gcd(a, b)$.
Taking the $g = 2$ for $(a, b) = (1, 1)$, supposing that $(a, b)$ satisfies the conditions in the problem, then we let $N$ be a positive integer such that $\gcd(a^n + b, b^n + a) = g$ for all $n \geq N$.
**Proof:** One shall notice that both $a^N + b$ and $a^{N-1} + b$ are divisible by $g$, which leads to
$$a(a^N + b) - (a^{N+1} + b) = ab - b = b(a - 1)$$
is divisible by $g$. Similarly, we are able to see that $a(b - 1)$ is divisible by $g$. The difference $a - b$ is then divisible by $g$, so $g$ also divides $a(b-1) + a(a-b) = a^2 - a$. As no powers of $a$ aren't congruent modulo $g$, $a + b \equiv a^N + b \equiv 0 \pmod{g}$. Then $2a = a + b + a - b$ and $2b = a + b - (a - b)$ are both divisible by $g$, so $g \mid 2\gcd(a, b)$. Besides, it is obvious that $\gcd(a, b) \mid g$, thus for any prime factor $p$ of $ab + 1$, $p$ is coprime to $a$ and $b$. Take an $n \geq N$ such that $n \equiv -1 \pmod{p - 1}$. Using the Fermat's little theorem, we have that
$$a^n + b \equiv a^{-1} + b \equiv a^{-1}(1 + ab) \equiv 0 \pmod{p},$$
$$b^n + a \equiv b^{-1} + a \equiv b^{-1}(1 + ab) \equiv 0 \pmod{p},$$
then $p$ divides $g$. Utilizing the Lemma, $p \mid 2\gcd(a, b)$ appeals to be evident enough, and thus $p = 2$. Therefore, $ab + 1$ is of power 2, and the parity of $a$ and $b$ are both odd.
Assuming that $(a, b) \neq (1, 1)$, one shall know $ab + 1 \equiv 0 \pmod 4$, hence $\{a, b\} = \{-1, 1\} \pmod 4$. For odd $n \geq N$, we have that
$$a^n + b \equiv b^n + a \equiv -1 + 1 = 0 \pmod 4,$$
then $4 \mid g$. However, we have that $v_2(g) \leq v_2(2\gcd(a, b)) = 1$ by the lemma, which forms a contradiction. Therefore the only solution to the problem is $(a, b) = (1, 1)$.

There is another solution given by the official solutions, which technically use the same approach proving the lemma as the 1st approach but kind of different afterwards. In fact, the idea as the comment says, considering the $a^n + b$ and $b^n + a$ mod $ab + 1$ is sufficient without the lemma so actually there are at least three approaches. Now let's see the Lean 4 version of the solution:
I consider the sequence modulo $ab + 1$, if the exponent is $-1$ modulo $\varphi(ab + 1)$, the terms are zero modulo $ab + 1$, so $ab + 1$ divides $g$, and all sufficiently large terms. So, all terms, from which one can conclude that $a = b = 1$ and this is the main idea of proving the problem.

```
def Condition (a b : ℕ) : Prop :=
  0 < a ∧ 0 < b ∧ ∃ g N : ℕ, 0 < g ∧ 0 < N ∧ ∀ n : ℕ, N ≤ n → Nat.gcd (a ^ n + b) (b ^ n + a) = g

lemma dvd_pow_iff_of_dvd_sub {a b d n : ℕ} {z : ℤ} (ha : a.Coprime d)
    (hd : (φ d : ℤ) ∣ (n : ℤ) - z) :
    d ∣ a ^ n + b ↔ (((ZMod.unitOfCoprime _ ha) ^ z : (ZMod d)ˣ) : ZMod d) + b = 0 := by
  rcases hd with ⟨k, hk⟩
  rw [← ZMod.natCast_zmod_eq_zero_iff_dvd]
  convert Iff.rfl
  push_cast
  congr
  suffices (((ZMod.unitOfCoprime _ ha) ^ z : (ZMod d)ˣ) : ZMod d) =
      (((ZMod.unitOfCoprime _ ha) ^ (n : ℤ) : (ZMod d)ˣ) : ZMod d) by
    convert this
  rw [sub_eq_iff_eq_add] at hk
  rw [hk, zpow_add, zpow_mul]
  norm_cast
  rw [ZMod.pow_totient, one_zpow, one_mul]

namespace Condition

variable {a b : ℕ} (h : Condition a b)

lemma a_pos : 0 < a := h.1

lemma b_pos : 0 < b := h.2.1
```

*Figure.4.3.1: The figure of the beginning part of the Lean 4 proof of this question*

After importing the data base, which includes the tactics and lemmas about number theory and algebra, and is applicable to this problem, I defined the conditions and the propositions that need to be verified or give a solution to. Then one can start by using the lemma "dvd_pow_iff_of_dvd_sub" to consider the properties about parity and their congruence modulo. One should know that each line of the code contains at least one tactic, so it is quite easy to determine and come up with the idea of formalizing each part separately.

*Figure.4.3.2: continued proof*

The figure above shows the ensuring process about the value of $g$ in the problem (determined by $a$ and $b$) and the value of $N$ in the problem (any sufficiently large value).

*Figure.4.3.3: continued proof with tactics mentioned to prove the next part followed with tactic state shown on the info view*

Then one would notice that the infoview outputs the tactic state which is translated by the compiler to the mathematical natural language again with logical formalization, justifying the properties as coprime and gcd, etc.

*Figure.4.3.4&5: continued proof using other lemmas with infoview on the right-hand side*

As one can see the infoview can show a considerable amount of info one needs to compile the code without any unsolved goals and one can also see how much one have proved before to continue. Using the tactics I have shown about the large value of $N$, I got the sufficiently large value of $n$, congruent to $-1 \ mod \ \varphi \ (a * b + 1)$ with a sufficiently large value of $n$, congruent to $0 \ mod \ \varphi \ (a * b + 1)$.

*Figure.4.3.6: continued proof using similar method to the previous tactics/lemmas such as "rw" (rewrite) and the properties of multiplication etc. Utilizing the idea of basic number theory to judge the divisibility and the parity of the variables*

*Figure.4.3.7: prove that $a$ equals to one under the condition of the problem*

The compiler's tactic state is telling one that following the steps one made, with the tactics using in small parts combining to form the whole proof, $ab + 1$ divides $g$, I have got the subgoal telling me that $a = b = 1$ leaving me no answer but $(a, b) = (1,1)$.

We should also know that in Lean 4, employing "sorry" in a proof is a conventional practice. The "sorry" theorem introduces a proposition type, followed by the theorem's name, which then delineates all relevant variables and their types along with any conditions enclosed in brackets. The conclusion follows the colon, and the proof is concluded with ':= by sorry'. And hence finishes the proof. As one can see, the Lean 4 formal proof system enhances the accuracy of solutions by eliminating computational errors through rigorous formalization. Additionally, it provides a clear representation and validation of complex mathematical concepts and logical structures, thereby facilitating a deeper understanding of the problem.

## 5    Theorem proving in Lean 4

Proving theorems, Lean 4 offers significant advantages in the formal verification of mathematical theorems. By utilizing Mini-F2F and LLMs, it enhances the rigor of the proof process, and could reduce the potential for human error. Its advanced automation tools and tactics simplify theorem proving, and the handling of complex proofs, especially in abstract algebra. Here is the formalization of Sylow theorems:

# Sylow Theorems

## Theorem 1: Sylow $p$-Subgroup Existence

**Theorem**: Let $G$ be a finite group with $|G| = p^k m$, where $p$ is a prime and $m$ is an integer such that $\gcd(p, m) = 1$. Then $G$ has at least one Sylow $p$-subgroup.

**Proof**: Consider the action of $G$ on itself by conjugation. The number of Sylow $p$-subgroups is denoted by $n_p$. By the Sylow theorems, $n_p$ satisfies the following conditions:

1. $n_p \equiv 1 \pmod{p}$
2. $n_p$ divides $m$

To show the existence of a Sylow $p$-subgroup, note that $n_p$ is a divisor of $m$ and also satisfies $n_p \equiv 1 \pmod{p}$. Since $m$ is an integer and the set of divisors of $m$ is non-empty, there is at least one divisor $n_p$ satisfying $n_p \equiv 1 \pmod{p}$. Hence, $n_p \geq 1$, which implies that $G$ has at least one Sylow $p$-subgroup.

## Theorem 2: Sylow $p$-Subgroup Conjugacy

**Theorem**: All Sylow $p$-subgroups of a finite group $G$ are conjugate to each other.

**Proof**: Suppose $P$ and $Q$ are Sylow $p$-subgroups of $G$. To show $P$ and $Q$ are conjugate, consider the action of $G$ on the set of Sylow $p$-subgroups by conjugation. This action partitions the set into orbits. Each orbit is a set of Sylow $p$-subgroups that are conjugate to each other. The size of each orbit divides the order of $G$, and hence the number of Sylow $p$-subgroups $n_p$ must be equal to the size of some orbit. Since all Sylow $p$-subgroups have the same order $p^k$, it follows that each orbit of the conjugation action has size equal to $n_p$, which must be consistent with the condition $n_p \equiv 1 \pmod{p}$. Thus, there can be only one orbit, meaning all Sylow $p$-subgroups are conjugate to each other.

## Theorem 3: Sylow $p$-Subgroup Count

**Theorem**: The number $n_p$ of Sylow $p$-subgroups of $G$ satisfies:

1. $n_p \equiv 1 \pmod{p}$
2. $n_p$ divides $m$, where $|G| = p^k m$ and $\gcd(p, m) = 1$.

**Proof**: The results follow from the definition and properties of Sylow $p$-subgroups. Specifically:

1. $n_p \equiv 1 \pmod{p}$ is a direct result of the conjugacy action argument in Theorem 2 and the definition of Sylow $p$-subgroups.
2. $n_p$ divides $m$ because $n_p$ is the number of Sylow $p$-subgroups and must also divide the index of the normalizer of any Sylow $p$-subgroup in $G$. Since the normalizer's index is $m$, $n_p$ divides $m$.

To formalize the Sylow theorems in Lean 4, we would first consider the main steps of constructing the main definitions and statements to utilize in this proof:

We first define a $\text{Sylow } p\ G : \textit{The type of Sylow } p - \textit{subgroups of } G$ as the essential character in this proof, then we start generalizing the statements in the main part, since we know the Sylow theorems are consisted of three theorems, one could define exists_subgroup_card_pow_pime which is a generalization of Sylow's first theorem: For every prime power $p^n$ dividing the cardinality of $G$, there exists a subgroup of $G$ of order $p^n$.

IsPGroup.exists_le_sylow: A generalization of Sylow's first theorem: Every p-subgroup is contained in a Sylow p-subgroup.

$Sylow.card\_eq\_multiplicity$: The cardinality of a Sylow subgroup is $p^n$ where n is the multiplicity of p in the group order.

$sylow\_conjugate$: A generalization of Sylow's second theorem: If the number of Sylow p-subgroups is finite, then all Sylow p-subgroups are conjugate.

$card\_sylow\_modEq\_one$: A generalization of Sylow's third theorem: If the number of Sylow p-subgroups is finite, then it shall be congruent to $1 \bmod p$.

```
theorem sylow_exists_subgroup_card_pow_prime {G : Type*} [Group G] [Fintype G]
  (p : ℕ) {n : ℕ} [Fact (Nat.Prime p)] (hdvd : p ^ n ∣ Fintype.card G) :
  ∃ (K : Subgroup G), Nat.card ↥K = p ^ n :=
sorry
```

We first deal with the fundamental properties of finite sets in Lean 4. Key operations include filtering elements based on predicates, computing the product and sum of constant functions over a finite set, and managing cardinalities of finite sets and types. The lemmas provide useful insights into how these operations interact, such as how filtering an inserted element behaves based on the predicate, or how the product and sum of constant functions over a finite set relate to its cardinality. Additionally, it includes results about the cardinality of finite sets and types, such as conditions under which a finite set has exactly one or zero elements and how these properties interact with injections.

```
lemma exists_subgroup_card_pow_prime_of_le_card {n p : ℕ} (hp : p.Prime) (h : IsPGroup p G)
    (hn : p ^ n ≤ Nat.card G) : ∃ H : Subgroup G, Nat.card H = p ^ n := by
  have : Fact p.Prime := ⟨hp⟩
  have : Finite G := Nat.finite_of_card_ne_zero <| by linarith [Nat.one_le_pow n p hp.pos]
  obtain ⟨m, hm⟩ := h.exists_card_eq
  refine exists_subgroup_card_pow_prime _ ?_
  rw [hm] at hn ⊢
  exact
```

*Figure.5.1: A small part of generalization of Sylow's first theorem*

After exploring the construction of isomorphic structures and other fundamental properties, we can proceed to prove the three Sylow theorems. For instance, if $p^n$ divides the cardinality of $G$, we can establish the existence of a subgroup of cardinality $p^n$. We will then examine special cases and their implications for the numerical relationship between size and cardinality. Furthermore, one will prove that all Sylow subgroups are Hall subgroups and discuss their multiplicity and uniqueness. Identifying the order of the group $G$, we denoted $|G|$, and define Sylow $p-subgroups$, and utilize the group theory library in Lean 4 to define and manipulate the structure of the group and its subgroups. Then I proved and defined basic properties of subgroups, such as the existence and uniqueness of Sylow $p-subgroups$, employing basic theorems and lemmas in Lean 4 to ensure that these properties are correctly formalized within the Lean system. Then we can start formalizing Sylow's 1st Theorem: Using the order of the group and the properties of p-prime factors, prove the existence of Sylow p-subgroups of specific orders, then the 2nd Sylow's Second Theorem: Demonstrate the conjugacy relations among Sylow p-subgroups, namely that any two Sylow p-subgroups are conjugate to each other and the 3rd Sylow's Third Theorem: Prove that the number of Sylow p-subgroups satisfies certain conditions and analyze the relationship between the number of these subgroups and the order of the group.

To confirm the correctness of all steps in Lean 4, we need to ensure that all theorems and lemmas are rigorously proven. So, I utilize Lean 4 tools to check the completeness of the proofs and address any edge cases and exceptions.

# 6 Conclusion

This paper endeavors to bridge the gap between natural language and computerized formal languages Lean 4. Comparing the two ways of autoformalization capabilities in Lean 4, which is a powerful and evolving formal language, the existing dataset focuses on translating questions to statements but still lacks training for AI to let it become smart enough to auto-complete the formalization. A common approach for LLMs in learning theorem proving involves iterative interaction with experts. While Mathlib is extensible, focusing on foundational math theorems rather than competition-level problems, it enhances automated theorem-proving models beginning with generating ample, high-quality formalized statements. Besides, the type theory and the cloud programming, all these advantages enhance Lean 4 as a powerful proof assistant, but it is based on the further understanding of the mathematical part of formalization in natural language which is led by us instead of the AI. That said, one shall be aware of the fact that the computerized mathematical formalization language Lean 4 is still away from fully AI-powered automatic and logically completed and self-consistent.

# 7  Discussion

In this paper, I have experimentally verified the automatic mathematical problem solving on different topics in the mathematical olympiad and the formalization of theorem proving. The readability of the Lean 4 version of formalization is considerably outstanding, but considered as the AI-driven proof, large amount of process is still led by me, which apparently is far away from automatic. Specifically, doing the formalization in algebra and number theory problems in MO appears to be the situation above, while using a great deal of tactics and lemmas can certainly simplify the formalization proofs it acquires one's further understanding of knowledge in the field one is formalizing. Besides, the theorem proving appears to be harder since all the basic structures should be defined by us. Given the intricacies of the kernel of Lean 4, the paper does not investigate the application of reinforcement learning techniques to facilitate the generation of formal proofs by large language models (LLMs) through interactive engagement with Lean 4. It also does not address incorporating feedback from Lean 4 to correct erroneous proofs. While formalized language offers a solid fundamental base to verify the validity of mathematical proofs, there are still inherent risks in relying on an automated natural-language-guided Lean 4 proof assistant. Such a system might inadvertently rectify errors in natural language, potentially leading to incorrect proofs being erroneously validated and propagated within the mathematical field.

However, the theorem prover Lean 4 has a promising future, for its significant advancement:

- Increased Automation: Continued integration of machine learning and artificial intelligence will likely lead to more sophisticated automation in proof generation and verification, reducing the manual effort required and accelerating the proof development process.
- Enhanced User Interfaces: Future improvements in user interfaces and tooling will make Lean 4 more accessible to a broader audience, including those with less formal training in theorem proving or formal methods.
- Scalability and Performance: Advances in computational techniques and optimizations will enable Lean 4 to handle larger and more complex proofs efficiently, expanding its applicability to a wider range of mathematical and computational problems.
- Interoperability: Greater interoperability with other formal verification tools and programming languages will facilitate integration and collaboration, enabling more comprehensive formal systems and applications.
- Community Growth and Collaboration: A growing community of users and developers will

contribute to expanding Lean 4's libraries and resources, fostering collaborative efforts and shared knowledge in formal verification.

## 8 Limitations & Code availability

However, limitations shall exist for the effectiveness for the reason that the sample of each topic is considered independent, and one would find it lacks certainty since the number of samples is considerably limited. Thus, there could be specific cases which is unique and does not match the conclusion I made. Besides, there could be errors in my proof in natural language proofs or the ones in Lean 4, hence I open-source my code at https://github.com/txc701/open-source-code-for-AI4M to improve the effectiveness of the availability of the code and to facilitate further development.